\RequirePackage{fix-cm}
\documentclass[smallextended]{svjour3}       
\smartqed  
\usepackage{algorithm}
\usepackage{algorithmicx,algpseudocode}
\usepackage{cite}
\usepackage{hyperref}
\usepackage{color}
\usepackage{amsmath,amssymb}
\usepackage{subcaption}
\usepackage{caption}
\usepackage{float}
\usepackage{lastpage,fancyhdr,graphicx}

\journalname{International Conference on Soft Computing in Data Science}
\begin{document}

\title{Multi-step Time Series Forecasting Using Ridge Polynomial Neural Network with Error-Output Feed-backs \thanks{\color{blue} This is a pre-peer-review version of an article published in International Conference on Soft Computing in Data Science. The final authenticated version is available online at: \url{http://link.springer.com/chapter/10.1007/978-981-10-2777-2_5}
}
}


\author{Waddah Waheeb         \and
        Rozaida Ghazali
}


\institute{Waddah Waheeb \at
              Faculty of Computer Science and Information Technology, Universiti Tun Hussein
Onn Malaysia, Batu Pahat, Johor, Malaysia \\
Computer Science Department, Hodeidah University, P.O. Box 3114 Alduraihimi, Hodeidah, Yemen\\
              \email{waddah.waheeb@gmail.com}           
           \and
           Rozaida Ghazali \at
              Faculty of Computer Science and Information Technology, Universiti Tun Hussein
Onn Malaysia, Batu Pahat, Johor, Malaysia\\
\email{rozaida@uthm.edu.my}
}


\maketitle

\begin{abstract}
Time series forecasting gets much attention due to its impact on many practical applications. Higher-order neural network with recurrent feedback is a powerful technique which used successfully for forecasting. It maintains fast learning and the ability to learn the dynamics of the series over time. For that, in this paper, we propose a novel model which is called Ridge Polynomial Neural Network with Error-Output Feedbacks (RPNN-EOFs) that combines the properties of higher order and error-output feedbacks. The well-known Mackey–Glass time series is used to test the forecasting capability of RPNN-EOFS. Simulation results showed that the proposed RPNN-EOFs provides better understanding for the Mackey–Glass time series with root mean square error equal to 0.00416. This re-sult is smaller than other models in the literature. Therefore, we can conclude that the RPNN-EOFs can be applied successfully for time series forecasting.


\end{abstract}

\section{Introduction}
\label{intro}

Time series forecasting approaches have been widely applied to many fields such as financial forecasting, weather forecasting, traffic forecasting, etc. The aim of time series forecasting is building an approach that use past observations to forecast the future. For example, using a series of data $x_{t-n},…,x_{t-2}, x_{t-1}, x_{t}$ to forecasts data values $x_{t+1},…, x_{t+m}$. Generally, time series forecasting approaches can be classified into two approaches; statistical-based and intelligent-based approaches. Due to the nonlinear nature of most of time series signals, intelligent-based approaches have shown better performance than statistical approaches in time series forecasting [1].

Artificial Neural network (ANN), which is inspired by biological nervous systems, is an example of intelligent-based approaches. ANN can learn from historical data and learn its weight matrices to construct model that can forecast the future. Due to the nonlinear nature and the ability to produce complex nonlinear input-output mapping, ANN have been used successfully for time series forecasting [2].

Generally, ANNs can be grouped into two groups based on network structure; feedforward and recurrent networks [2]. In feedforward networks, the data flows in one direction only from the input nodes to the output nodes through network connec-tions (i.e. weights). On other hand, the connections between the nodes in recurrent networks form a cycle.

Multilayer perceptron (MLP) is one of the most used feedforward ANNs in fore-casting tasks [3]. However, due to the multilayered structure of MLP, it needs a large number of units to solve complex nonlinear mapping problems, which results in low learning rate and poor generalization [4]. To overcome these drawbacks, different types of single layer higher order neural networks (HONNs) with product neurons were introduced. Ridge Polynomial Neural Network (RPNN) [5] is a feedforward HONNs that maintain fast learning and powerful mapping properties, and is thus suitable for solving complex problems [3].

Two recurrent versions of RPNNs are existed namely; the Dynamic Ridge Poly-nomial Neural Networks (DRPNN) [6] and Ridge Polynomial Neural Networks with Error Feedback (RPNN-EF) [7]. DRPNN uses the output value from the output layer as a feedback connection to the input layer. On other hand, RPNN-EF uses the net-work error which is calculated by subtracting the desired value from the forecast val-ue. The idea behind recurrent networks is learning the network the dynamics of the series over time. As a result, the network should use this memory when forecasting [8]. DRPNN and RPNN-EF have been successfully used for time series forecasting [1, 6, 7, 9].

Due to the success of DRPNN and RPNN-EF, in this paper we propose a model that combine the properties of RPNNs and output-error feedbacks recurrent neural networks. This model is called Ridge Polynomial Neural Network with Error-Output Feedbacks (RPNN-EOFs). We applied the RPNN-EOFs to the chaotic Mackey-Glass differential delay equation series which is recognized as a benchmark problem that has been used and reported by many researchers for comparing the generalization ability of different models [10-18].

This paper consists of six sections. Section 2 introduces the basic concepts of RPNN, DRPNN and RPNN-EF. We describe the proposed model in Section 3. Section 4 covers the experimental settings. Section 5 is about results and discussion. And finally, Section 6 concludes the paper.

\section{The Existing Ridge Polynomial Neural Network Based Models}
\label{related}
This section discusses the existing Ridge Polynomial Neural Networks based models, namely Ridge Polynomial Neural Networks (RPNNs), Dynamic Ridge Polynomial Neural Networks (DRPNNs) and Ridge Polynomial Neural Networks with Error Feed-back (RPNN-EF).

\subsection{Ridge Polynomial Neural Networks(RPNNs)}
RPNNs [5] are an example of feedforward higher order neural networks (HONNs) that use one layer of trainable weights. RPNNs maintain powerful mapping capabili-ties and fast learning properties of single layer HONNs [3]. They are constructed by adding different degrees of Pi-Sigma Neural Networks (PSNNs) blocks [19] until a defined goal is achieved. They can approximate any continuous function on a com-pact set in multidimensional input space with arbitrary degree of accuracy [5]. RPNNs utilize univariate polynomials which help to avoid an explosion of free pa-rameters that found in some types of higher order feedforward neural networks [5].

\subsection{Dynamic Ridge Polynomial Neural Networks(DRPNNs)}
DRPNNs [6] are one type of recurrent version of RPNNs. DRPNNs take advantage of network output value as an additional input to the input layer. They are provided with memories that help to retain information to be used later [6]. DRPNN is trained by the real time recurrent learning algorithm (RTRL) [20].

DRPNNs are more suitable than RPNNs for time series forecasting due the fact that the behavior of some time series signals related to some past inputs on which the present inputs depends. Interested readers for the application of DRPNNs for time series forecasting nay be referred to [1, 6, 9].

\subsection{Ridge Polynomial Neural Network with Error Feedback (RPNN-EF)}
Another recurrent type of RPNN is RPNN-EF [7] as shown in Fig. 1. Unlike DRPNN, RPNN-EF takes advantage of network error value as an additional input to the input layer. This error is calculated by taking the difference between the desired output and network output. Such error feedback is also used in the literature with Functional Link Network (FLN) and Adaptive Neuro-Fuzzy Inference System (ANFIS) model [21, 22]. Like DRPNN, RPNN-EF is trained by RTRL algorithm.

RPNN-EF showed better understanding for multi-step ahead forecasting than RPNN and DRPNN. Furthermore, RPNN-EF was significantly faster than other RPNN-based models for one-step ahead forecasting [7].

\begin{figure}
\includegraphics[width=0.95\textwidth]{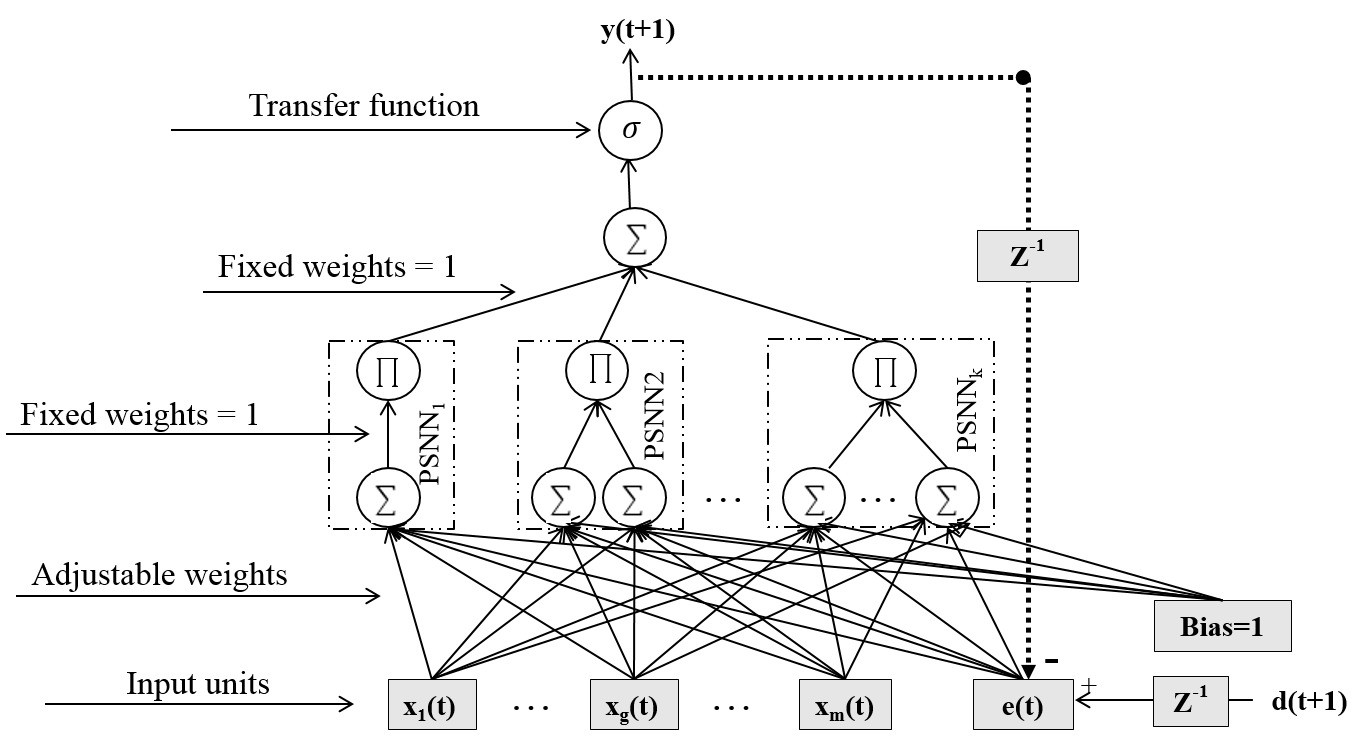}
\caption{Ridge Polynomial Neural Networks with Error Feedback. PSNN, d(t) stands for Pi-Sigma Neural Network and the desired output at time t.}
\end{figure}

\section{The Proposed Model: Ridge Polynomial Neural Network with Error-Output Feedbacks (RPNN-EOF)}
Due to the success of DRPNN, RPNN-EF for time series forecasting [1, 6, 7, 9], we propose Ridge Polynomial Neural Networks with Error-Output Feedbacks (RPNN- EOFs). This model combines the properties of RPNNs, and the powerful of both error and output feedbacks.

Generic network architecture of the RPNN-EOFs using Pi-Sigma neural networks as basic building blocks is shown in Fig. 2. Like other RPNN based models, RPNN-EOFs uses constructive learning method. That means the network structure grows from a small network and the network becomes larger as learning proceeds until the desired level of defined error is reached [5].

\begin{figure}
\includegraphics[width=0.95\textwidth]{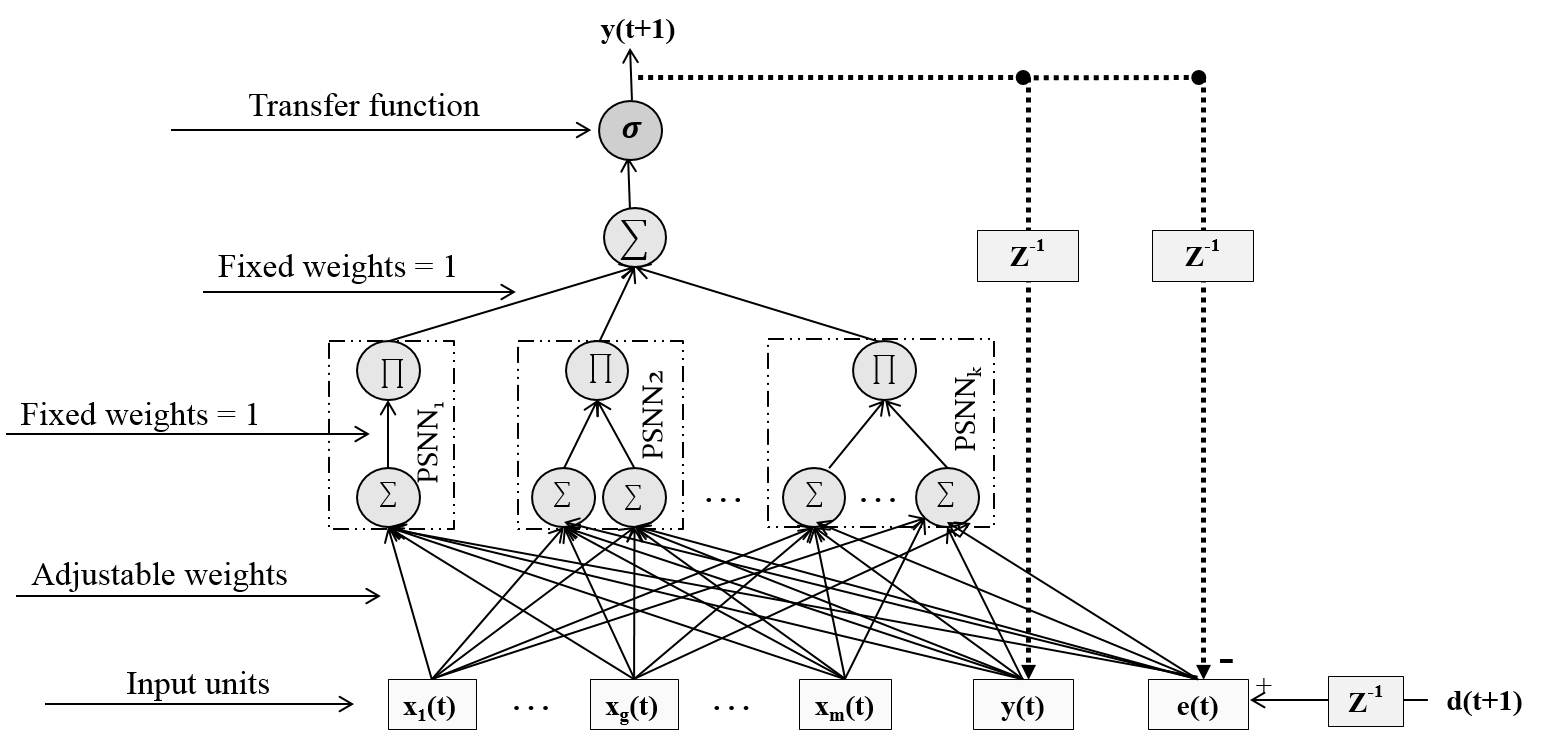}
\caption{Ridge Polynomial Neural Networks with Error-Output Feedbacks. PSNN, d(t) stands for Pi-Sigma Neural Network and the desired output at time t. Bias nodes are not shown here for reason of simplicity.}
\label{fig:1}       
\end{figure}

RPNN-EOFs is trained by RTRL algorithm. The output of RPNN-EOFs, which is denoted by y(t), is calculated as follows:
\begin{subequations}
\label{eq:erpnn02}
\begin{eqnarray}
y(t)\thickapprox\sigma\left(\sum_{i=1}^{k}P_{i}(t)\right)\\
P_{i}(t)=\prod_{j=1}^{i}\left(h_{j}(t)\right)\\
h_{j}(t)=\sum_{g=1}^{M+2}(w_{gj}*Z_{g}(t))+w_{0j}
\end{eqnarray}
\end{subequations}
where $P_{i}(t)$ is the output of Pi-Sigma block, $\sigma$ is the transfer function, $h_{j}(t)$ is the net sum of the sigma unit $j$, $w_{0j}$ is the bias, $w_{gj}$ is the weights between input and sigma units, and $Z(t)$ is the inputs which given as follow:

\begin{eqnarray}
\label{eq:erpnn01}
Z_{g}(t)=\begin{cases}
x_{g}(t) & 1\leq g\leq M\\
e(t-1)=d(t-1)-y(t-1) & g=M+1\\
y(t-1) & g=M+2
\end{cases}
\end{eqnarray}

Network error is calculated using the sum squared error as follows:

\begin{equation}
\label{eq:erpnn03}
E(t)=\frac{1}{2}\sum_{}^{}e(t)^{2}
\end{equation}

\begin{equation}
\label{eq:erpnn04}
e(t)=d(t)-y(t)
\end{equation}
where $d$ is the desired output and $y$ is the predicted output. At every time $t$, the weights changes are calculated as follows:

\begin{equation}
\label{eq:erpnn05}
\triangle w_{gl}=-\eta*\left(\frac{\partial E(t)}{\partial w_{gl}}\right)
\end{equation}
where $\eta$ is the learning rate. The value of $\frac{\partial E(t)}{\partial w_{gl}}$ is determined as:

\begin{equation}
\label{eq:erpnn06}
\frac{\partial E(t)}{\partial w_{gl}}= e(t)*\frac{\partial e(t)}{\partial w_{gl}}
\end{equation}

\begin{equation}
\label{eq:erpnn066}
\frac{\partial e(t)}{\partial w_{gl}}=\frac{\partial e(t)}{\partial y(t)}*\frac{\partial y(t)}{\partial w_{gl}}
\end{equation}

\begin{equation}
\label{eq:erpnn0661}
\frac{\partial e(t)}{\partial w_{gl}}=-\frac{\partial y(t)}{\partial w_{gl}}
\end{equation}

\begin{equation}
\label{eq:erpnn0662}
\frac{\partial y(t)}{\partial w_{gl}}=\frac{\partial y(t)}{\partial P_{i}(t)}*\frac{\partial P_{i}(t)}{\partial w_{gl}}
\end{equation}

From Equation (1), we have
\begin{equation}
\label{eq:erpnn066}
\frac{\partial y(t)}{\partial w_{gl}}=\frac{\partial y(t)}{\partial P_{i}(t)}*\frac{\partial P_{i}(t)}{\partial w_{gl}}= (y(t))^{'}*\left(\prod_{j=1,j\neq l}^{i}h_{j}(t)\right)*\left(Z_{g}(t) + \left(w_{(M+1)j}*\frac{\partial e(t-1)}{\partial w_{gl}}\right) + \left(w_{(M+2)j}* \frac{\partial y(t-1)}{\partial w_{gl}} \right) \right)
\end{equation}

Assume $D^{Y}$ and $D^{E}$ as dynamic system variables [2], where $D^{Y}$ and $D^{E}$ are:
\begin{equation}
\label{eq:erpnn10}
D^{Y}_{gl}(t)=\frac{\partial y(t)}{\partial w_{gl}}
\end{equation}
\begin{equation}
\label{eq:erpnn101}
D^{E}_{gl}(t)=\frac{\partial e(t)}{\partial w_{gl}}
\end{equation}

Substituting Equation (11) and Equation (12) into Equation (10), we have
\begin{equation}
\label{eq:erpnn11}
\frac{\partial y(t)}{\partial w_{gl}}=(y(t))^{'}*\left(\prod_{j=1,j\neq l}^{i}h_{j}(t)\right)*\left(Z_{g}(t) + \left(w_{(M+1)j}*D^{E}_{gl}(t-1)\right) + \left(w_{(M+2)j}* D^{Y}_{gl}(t-1) \right) \right)
\end{equation}

Recall Equation (8), from Equation (12), we have
\begin{equation}
\label{eq:erpnn102}
D^{E}_{gl}(t)=-D^{Y}_{gl}(t)
\end{equation}

Substituting Equation (14) into Equation (13), we have
\begin{equation}
\label{eq:erpnn11-1}
\frac{\partial y(t)}{\partial w_{gl}}=(y(t))^{'}*\left(\prod_{j=1,j\neq l}^{i}h_{j}(t)\right)*\left(Z_{g}(t) + D^{Y}_{gl}(t-1) \left(w_{(M+2)j}-w_{(M+1)j} \right) \right)
\end{equation}

For simplification, the initial values for $D^{Y}_{gl}(t)=0$, and $e(t)=y(t)=0.5$  to avoid zero value of $D^{Y}_{gl}(t)$. Then the weights updating rule is derived by substituting Equation (15) and Equation (8) into Equation (6), we have
\begin{equation}
\label{eq:erpnn06-9}
\frac{\partial E(t)}{\partial w_{gl}}= -e(t)*D^{Y}_{gl}(t)
\end{equation}

Then, substituting Equation (16) into Equation (5), we have
\begin{equation}
\label{eq:erpnn12}
\triangle w_{gl}=\eta*e(t)*D^{Y}_{gl}(t)
\end{equation}

\section{Experimental Design}

\subsection{Mackey-Glass Differential Delay Equation}
In this paper, we used the well known chaotic Mackey-Glass differential delay equa-tion series. This series is recognized as a benchmark problem that has been used and reported by many researchers for comparing the generalization ability of different models [10-18]. Mackey-Glass time series is given by the following delay differential equation:

\begin{equation}
\frac{dx}{dt}=\beta x(t)+\frac{\alpha x(t-\tau)}{1+x^{10}(t-\tau)}
\end{equation}
where $\tau$ is the time delay. We chose the following values for the variables $\alpha=0.2$, $\beta=-0.1$, $x(0)=1.2$, and $\tau=17$. With this setting the series produce chaotic behavior. One thousand data point were generated. The first 500 points of the series were used as a training sample, while the remaining 500 points were used as out-of-sample data. We used four input variables $x(t), x(t-6), x(t-12), x(t-18)$ to predict $x(t+6)$. All these settings were used for fair comparison with other studies in the literature [10-18].

\subsection{Data Preprocessing}
We scaled the points to the range [0.2, 0.8] to avoid getting network output too close to the two endpoints of sigmoid transfer function [1]. We used the minimum and maximum normalization method which is given by:
\begin{equation}
\dot{x}=(max_{2}-min_{2})*\left(\frac{x-min_{1}}{max_{1}-min_{1}}\right)+min_{2}\label{eq:norm}
\end{equation}
where $x$ refers to the observed (original) value, $\dot{x}$ is the normalized version of $x$, $min_{1}$, and $max_{1}$ are the respective minimum and maximum values of all observations, and $min_{2}$, and $max_{2}$ refer to the desired minimum and maximum of the new scaled series.

\subsection{Network Topology}
The topology of the RPNN-EOFs that we used is shown in Table 1. Most of the settings are either based on the previous works with RPNN based models that found in the literature [1, 6, 7, 9] or by trial and error.

\begin{table}[H]
\centering
\caption{ Network topology.}
\begin{tabular}{p{5cm}p{6.5cm}}
\hline\noalign{\smallskip}
Setting & Value \\
\noalign{\smallskip}\hline\noalign{\smallskip}
Activation function &  Sigmoid function\\ 
Number of Pi-Sigma block (PSNN)&  Incrementally grown from 1 to 5\\ 
Stopping criteria &  Maximum number of epochs =3000 or after accomplishing the 5th order network learning.\\ 
Initial weights&  [-0.5,0.5]\\ 
Momentum&  [0.4-0.8]\\ 
Learning rate ($\eta$)&  [0.01-1]\\ 
Decreasing factors for ($\eta$) &  0.8\\ 
Threshold of successive PSNN addition ($r$)&  [0.00001-0.1]\\ 
Decreasing factors for $r$&  [0.05, 0.2]\\
\noalign{\smallskip}\hline
\end{tabular}
\label{topology_training}
\end{table}

\subsection{Performance Metrics}
Because we aim to compare our proposed model’s performance with other models in the literature, we used the Root Mean Squared Error (RMSE) metric. RMSE is the standard metric wich used by many researchers with Mackey-Glass series [10-18]. The equation for RMSE is given by:

\begin{equation}
RMSE=\sqrt{\frac{1}{N}\sum_{i=1}^{N}(Y_{i}-\hat{Y}_{i})^{2}}
\end{equation}

\section{Results and Discussion}
\label{exp}
The forecasting model of Mackey-Glass time series is built via the experimental de-sign settings. The best out-of-sample data forecasting for RPNN-EOFs is shown in Fig. 3. It can be seen that the RPNN-EOFs can follow the dynamic behavior of the series precisely. To show the difference between the output of the RPNN-EOFs and the out-of-sample points, which is called forecasting error, we plot the forecasting error in Fig. 4.

\begin{figure}
\includegraphics[width=1.15\textwidth]{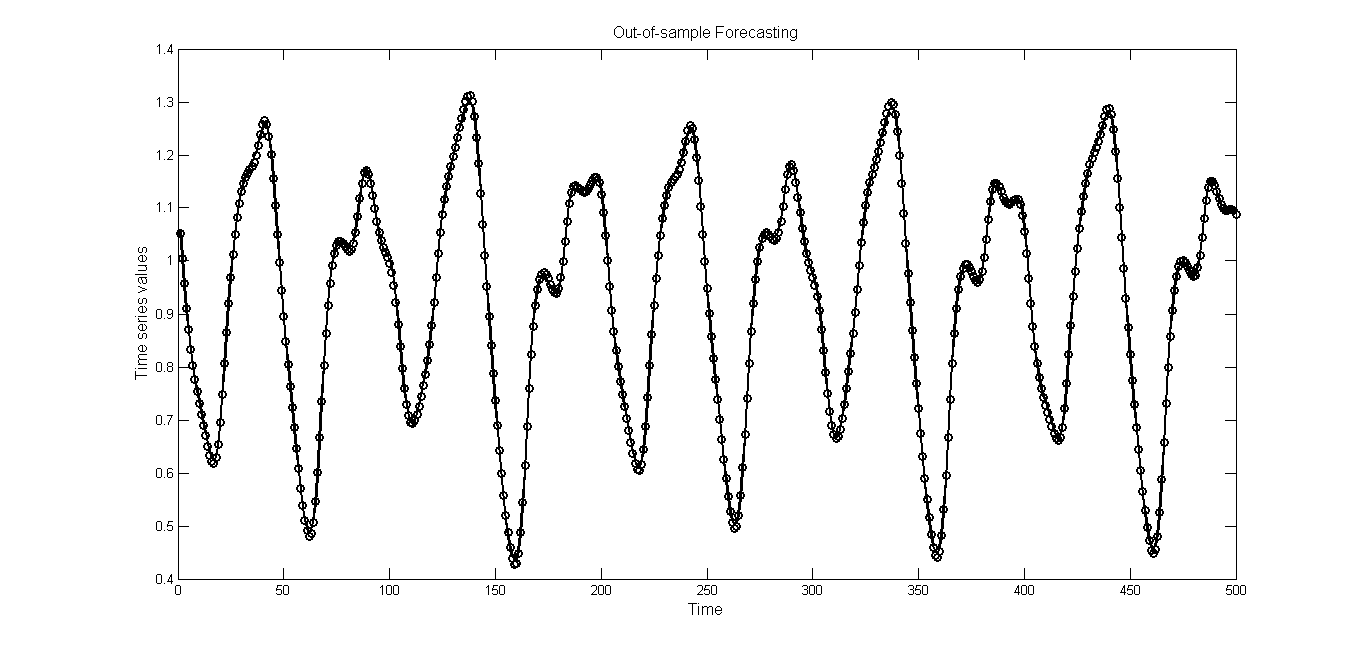}
\caption{The best forecasting results for Mackey-Glass time series using RPNN-EOF. Circle is the original series while the solid line is the forecast series.}
\label{fig:2}       
\end{figure}

\begin{figure}
\includegraphics[width=1.15\textwidth]{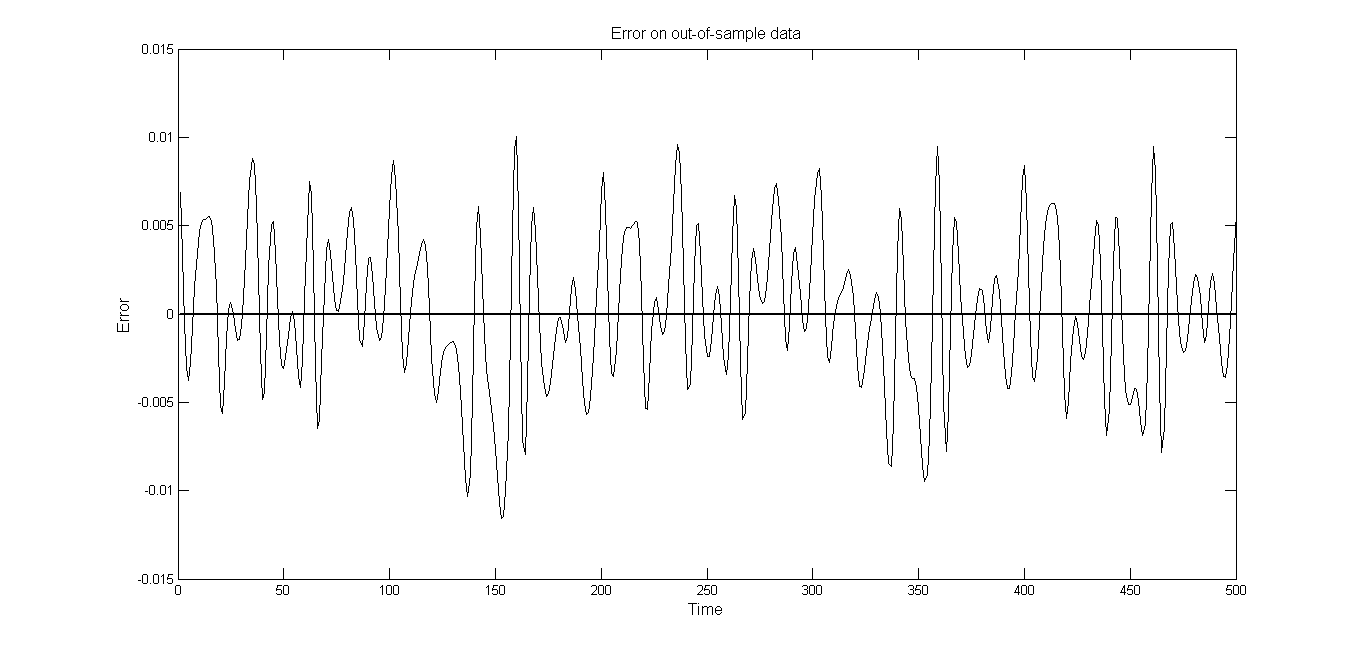}
\caption{Forecasting error for Mackey-Glass time series using RPNN-EOF.}
\label{fig:3}       
\end{figure}

For fair comparison with recent studies, we de-normalized the results of RPNN-EOFs model. Table 2 lists the generalization capabilities of other methods [10-18]. The generalization capabilities were measured by applying each model to forecast the out-of-sample data. The results show that the proposed RPNN-EOFs model offers a smaller RMSE than the other models. Based on these results, we can conclude that the RPNN-EOFs alone provides better understanding for the series and smaller error in comparison to other models.

\begin{table}[H]
\caption{Comparison of the performance of various existing models.}
\label{tab:4}       
\begin{tabular}{p{9.8cm}p{1.3cm}}
\hline\noalign{\smallskip}
Model & RMSE  \\
\noalign{\smallskip}\hline\noalign{\smallskip}
Differential evolution - beta basis function neural networks (DE-BBFNN) [10]&	0.030\\
Dynamic evolving computation system (DECS) [11]&	0.0289\\
Orthogonal function neural network [12]&	0.016\\
Multilayer feedforward neural network - Backpropagation algorithm (MLFBP) [13]&	0.0155\\
Backpropagation network optimized by hybrid K-means-greedy algorithm [14]&	0.015\\
Modified differential evolution and the radial basis function (MDE-RBF) [15]&	0.013\\
Functional-link-based neural fuzzy network optimized by hybrid of cooperative particle swarm optimization and cultural algorithm (FLNFN-CCPSO) [16]&	0.008274\\
Multilayer neural network with the multi-valued neurons- QR decomposition (MLMVN-QR) [13]&	0.0065\\
Wavelet neural network with hybrid learning approach (WNN-HLA) [17]&	0.006\\
Multilayer neural network with the multi-valued neurons (MLMVN) [13]&	0.0056\\
\textbf{RPNN-EOF (proposed)}&	\textbf{0.00416*}\\
Grid-based fuzzy systems with 192 rules [18]&	0.0041\\
Multigrid-based fuzzy system 3 sub-grids with 120 rules [18]&	0.0031\\
\noalign{\smallskip}\hline
\end{tabular}
\begin{flushleft} $^{*}$This is the de-normalized value for the RMSE.
\end{flushleft}
\end{table}

\section*{Conclusions and Future Works}
This paper investigated the forecasting capability of the Ridge Polynomial Neural Network with Error-Output Feedbacks (RPNN-EOFs) for multi-step time series fore-casting. The well-known Mackey–Glass differential delay equation was used to test the forecasting capability of RPNN-EOFs. Simulation results showed that the pro-posed RPNN-EOFs provides better understanding for the Mackey–Glass time series and smaller error in comparison to other models in the literature. Therefore, we can conclude that the RPNN-EOFs can be applied successfully for time series forecasting. The future works will be using more time series to ensure the good performance of the proposed model.

\begin{acknowledgements}
The authors would like to thank Universiti Tun Hussein Onn Malaysia (UTHM) and Ministry of Higher Education (MOHE) Malaysia for financially supporting this re-search under the Fundamental Research Grant Scheme (FRGS), Vote No. 1235.
\end{acknowledgements}


\section*{References}
1.	Al-Jumeily, D., Ghazali, R., Hussain, A.: Predicting Physical Time Series Using Dynamic Ridge Polynomial Neural Networks. PLOS ONE, (2014)

\noindent 2.  Haykin, S. S.: Neural Networks and Learning Machines. Prentice Hall, New Jersey (2009)

\noindent 3. Ghazali, R., Hussain, A.J., Liatsis, P., Tawfik, H.: The Application of Ridge Polynomial Neural Network to Multi-step Ahead Financial Time Series Prediction. Neural Comput. Appl. 17, no. 3, 311—323 (2008)

\noindent 4. Yu, X., Tang, L., Chen, Q., Xu, C.: Monotonicity and Convergence of Asynchronous Update Gradient Method for Ridge Polynomial Neural Network. Neurocomputing, 129, 437--444 (2014)

\noindent 5.  Shin, Y., Ghosh, J.: Ridge Polynomial Networks. IEEE T. Neural Networ. 6(3),  610—622 (1995)

\noindent 6. Ghazali, R., Hussain, A. J., Nawi, N.M., Mohamad, B.: Non-Stationary and Stationary Prediction of Financial Time Series Using Dynamic Ridge Polynomial Neural Network. Neurocomputing 72(10), 2359--2367 (2009)

\noindent 7.  Waheeb, W., Ghazali, R., Herawan, T.: Time Series Forecasting Using Ridge Polynomial Neural Network with Error Feedback. Proceedings of The Second International Conference on Soft Computing and Data Mining (SCDM-2016) (in press)

\noindent 8. Samarasinghe, S.: Neural Networks for Applied Sciences and Engineering: From  Fundamentals to Complex Pattern Recognition. CRC Press, New York (2006)

\noindent 9. Ghazali, R., Hussain, A. J., Liatsis, P.: Dynamic Ridge Polynomial Neural Network: Forecasting the univariate Non-stationary and Stationary Trading Signals. Expert Syst. Appl. 38(4), pp.3765--3776 (2011)

\noindent 10. Dhahri H, Alimi A.: Automatic selection for the beta basis function neural networks. In Nature Inspired Cooperative Strategies for Optimization (NICSO 2007), pp. 461-474, Springer Berlin Heidelberg (2008)

\noindent 11. Chen YM, Lin CT.: Dynamic parameter optimization of evolutionary computation for on-line prediction of time series with changing dynamics, APPL SOFT COMPUT, 31;7(4), 1170-6, (2007)

\noindent 12.Wang H, Gu H.: Prediction of chaotic time series based on neural network with Legendre polynomials. In International Symposium on Neural Networks, pp. 836-843, Springer Berlin Heidelberg, (2009)

\noindent 13. Aizenberg I, Luchetta A, Manetti S.: A modified learning algorithm for the multilayer neural network with multi-valued neurons based on the complex QR decomposition, Soft Comput., 1;16(4), 563-75, (2012)

\noindent 14. Tan J.Y., Bong D.B., Rigit A.R.: Time Series Prediction using Backpropagation Network Optimized by Hybrid K-means-Greedy Algorithm, Engineering Letters. 1;20(3), 203-10, (2012)

\noindent 15. Dhahri H, Alimi AM.: The modified differential evolution and the RBF (MDE-RBF) neural network for time series prediction. In the 2006 IEEE International Joint Conference on Neural Network Proceedings, pp. 2938-2943, IEEE (2006)

\noindent 16. Lin CJ, Chen CH, Lin CT.: A hybrid of cooperative particle swarm optimization and cultural algorithm for neural fuzzy networks and its prediction applications. IEEE Transactions on Systems, Man, and Cybernetics, Part C (Applications and Reviews), 39(1), 55-68, (2009)

\noindent 17. Lin CJ.: Wavelet neural networks with a hybrid learning approach. Journal of Information science and Engineering, 1;22(6):1367-87 (2006)

\noindent 18. Herrera LJ, Pomares H, Rojas I, Guillén A, González J, Awad M, Herrera A.: Multigrid-based fuzzy systems for time series prediction: CATS competition, Neurocomputing. 31;70(13), 2410-25, (2007)

\noindent 19. Shin, Y., Ghosh, J.: The Pi-Sigma Network: an Efficient Higher-Order Neural Network for Pattern Classification and Function Approximation. In Neural Networks, 1991., IJCNN-91-Seattle International Joint Conference, Vol. 1, pp. 13—18. IEEE. (1991)

\noindent 20. Williams, R.J., Zipser, D.: A Learning Algorithm for Continually Running Fully Recurrent Neural Networks. Neural. Comput. 1(2), 270--280 (1989)

\noindent 21. Dash PK, Satpathy HP, Liew AC, Rahman S.: A real-time short-term load forecasting system using functional link network. IEEE T POWER SYST, 12(2), 675-80, (1997)

\noindent 22. Mahmud, M.S., Meesad, P.: An Innovative Recurrent Error-Based Neuro-Fuzzy System with Momentum for Stock Price Prediction. Soft. Comput. 1--19 (2015)

%
%

\end{document}